\documentclass{article}

\usepackage{blindtext}
\usepackage{hyperref}
\usepackage{graphicx}
\usepackage[final]{pdfpages}
\usepackage{afterpage }
\usepackage{textcomp}
\usepackage{setspace}
\usepackage{authblk}
\usepackage{cite}
\usepackage{amsfonts}
\usepackage{algorithm}
\usepackage{enumitem}

\let\oldfootnote\footnote
\def\footnote{\ifhmode\unskip\fi\oldfootnote}

\usepackage{float}
\restylefloat{table}
 
\title{
    AlteregoNets: a way to human augmentation \\ 
    \author[1]{
        Dr. David Kupeev 
        \footnote{kupeev@gmail.com,  
        \copyright  2018 David Kupeev
        }  
    }
    \affil[1]{ 
    Rafael Advanced Defense Systems Ltd.,
    POB 2250, Haifa, 3102102 Israel}          
\date{}  
}

\providecommand{\keywords}[1]{\textit{Keywords:} #1}

\begin{document}

\maketitle

\begin{abstract}

A person dependent network,  called an AlterEgo net, is proposed for development.  The networks 
are created {\em per person}. It receives 
at input an object descriptions and 
outputs a simulation of the internal person's representation 
of the objects.

The network generates a textual stream  resembling the narrative stream of consciousness depicting multitudinous thoughts and feelings
{\em related to a perceived object}.
In this way, the 
object is described 
not by a 'static' set of its properties, like a dictionary,
but by 
the stream of words and word combinations referring to the object.

The network simulates a person's dialogue with a representation of the object. 
It
is based on an introduced algorithmic scheme, where 
perception is modeled by two interacting iterative cycles, reminding one  respectively the forward 
and backward propagation 
executed at training convolution neural networks. 
The 'forward' iterations generate a stream representing the 'internal world' of a human. The 'backward' iterations generate a stream representing an internal representation of the object. 

People perceive the world differently. Tuning AlterEgo nets to a specific person or group of persons, will allow simulation of their thoughts and feelings. 
Thereby these nets is potentially a new human augmentation technology for various applications.

\end{abstract}

\vspace*{0.5cm}


\tableofcontents

\section{Introduction}
\label{intro}  
Undoubtedly, modern computers perform arithmetical computations better than humans.
Also, in recent years, the computers began to solve the problems of detection and classification in images
in many hearings better than the man. 
Likewise, the topic of algoritmic simulation of human perception of objects received some coverage in the research literature\cite{lit1,lit2,lit3,lit4}.
However, to the best of our knowledge, to this day there have been no attempts to algorithmically {\em generate the context} that approximates what people  {\em perceive}. This article is such an attempt.

Our goal is to enable computers to yield data which approximates internal human representations of  perceived objects. 
These representation are the flows of thoughts \cite{thought} and
emotions \cite{feeling} related to an object, or groups of objects.

\begin{figure}
    \centering
    \includegraphics[scale=.55]{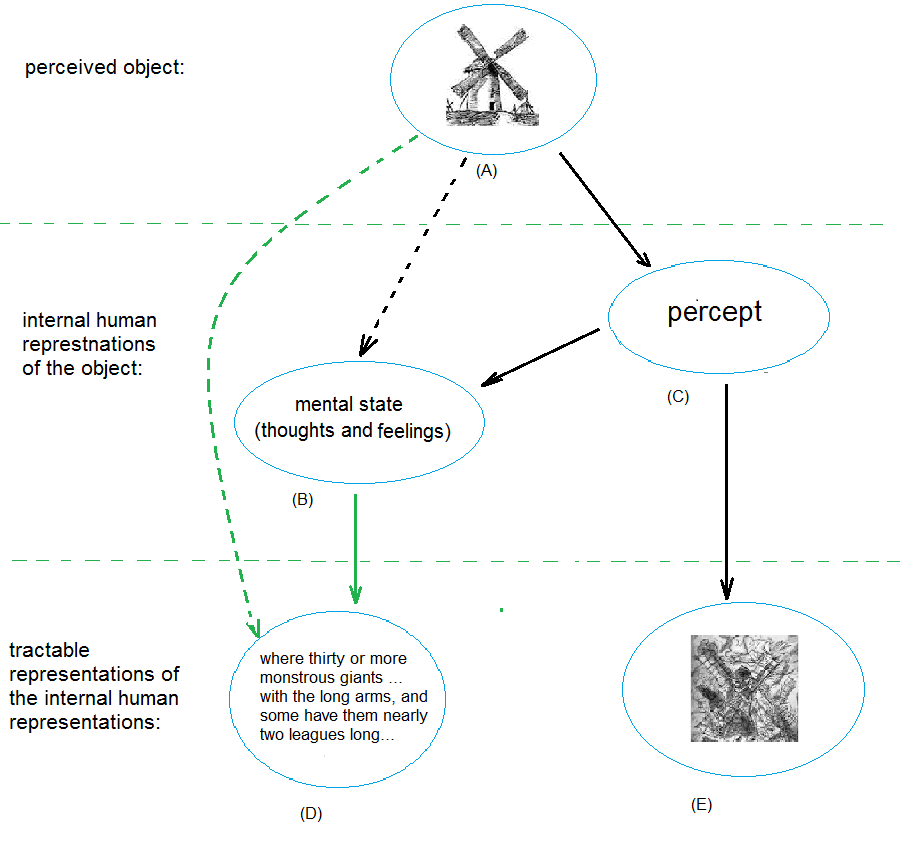}
    \caption{
     Relations between representations of  perceived objects.
    (A): The image \cite{commonDQ} perceived by  Don
     Quixote in the episode with windmills \cite{DQText}.  
    (B): The flow of thoughts \cite{streamP} modeled
    in this paper. 
    (C): The percept \cite{Percept} of the input image. 
    (D), (E): The textual and image media representing 
   the human representations.
    The flow of thoughts
    is naturally represented as a text, 
    and the percept as an image.
    (D) depicts the representations 
    constructed in this
    article. The dashed arrow (AB) denotes  
    partial forming of the mental state
    bypassing the percept. For example, 
    the feel of a color of a shoe\cite{Percept}
    may be obtained without forming the whole image
    of the shoe.     
    Green dashed arrow (AD) denotes optional
    use of the tools for
    image annotations \cite{3dparty}. 
  }
\label{fig:FIG_THOUGHTS_PERCEPT}
\end{figure}

The objects are given to us in some modality, usually visual or textual. For example, the object depicted in Fig.~\ref{fig:FIG_THOUGHTS_PERCEPT} (A) is visual.  
One may assume that when the object is perceived,
its certain representation acts 'inside a person'. 
These representations
can not be reproduced or operated directly. One may refer to them as the 
first level object representations.
The percept \cite{Percept} (ibid, (C))
may be considered as  the mental 
image
\cite{PerceptFarlex}  
of the object.
It is naturally that the majority of scientific investigations
of the mental {\em image} focus upon {\em visual} mental imagery \cite{Mental_image}. 
This is what the person sees in a perceived image.
On the other hand, we may
consider 
the flow of thoughts\,\cite{streamP} and feelings\cite{feeling} {\em related to the object}. 
(In some sense one may see this as a 
flow of perceived object properties.)
This is the internal human representation which
we want to simulate.

One may consider the 
task of further representing the 
first level representations by 
tractable data, 
visual or textual, with which  
algorithms can operate. It may be 
viewed as constructing the second level
object representations. 
It seems that natural representation of
the percept is the image, and of
the flow of thoughts is the textual stream.
In this paper we describe an 
algorithm for 
simulating
flows of thoughts
by textual streams, resembling 
the narrative 
stream of councioseness \cite{streamN}.

A key feature of our approach is that
we represent the perceived object properties
not just as static attributes associated with the 
object (e.g., an object may be 'large' or 'small'),
but in dynamic fashion -- as a textual stream.
And it's not surprising. Indeed,
at the first representation level,
we may treat the human thoughts
related to a perceived object as 
a kind of the 'general' flow of thoughts\,\cite{streamP}.
And the second level representation of this flow generated by our algorithm (Sect.\ref{BiWheel}) may be seen as
a particular case of the 
narrative stream of consciousness\cite{streamN}.

The objects whose perception is simulated in our paper
are paintings, historical emblems the person may mentally interact with, social scenes, etc. We do not impose a formal criterion for selecting such objects, one 
may just note that the richer is the set of the 
notions which may be associated with the object, the more suitable is the object for simulating the process of its perception under our model.

We do not impose strict assumptions
about the nature of the flow of the human thoughts and emotions
we want to simulate (referred at Fig.\ref{fig:FIG_THOUGHTS_PERCEPT} (B)). Indeed, one may suggest 
that this flow may be roughly expressed
by means of a natural language\cite{NatLang}, but definitely 
this is {\em not} a natural language.
On the other hand, our algorithm yields a stream of words and word combinations, similar to ibid (D). However, there is no contradiction  -- we {\em represent} the flow of some (unknown) modality by textual data.

Also, it may be noted  that eventually our goal is not to approximate {\em well} the human perception. Indeed, 
the computer calculates not like humans, the computational models of visual recognition\cite{Deep_learning} differ from that by 
humans\cite{Visual_perception}, etc.
Hence our goal
is defined as a rough computer modeling of certain mechanisms of human perception.

This underlines a 'target gap': on the one hand we want to simulate the perception as best as possible,  and on the other we assume in advance that it can not be done perfectly. This raises the problem of validation of our simulation. What does a well-built simulation mean?
For example, in the research on image classification\cite{dbs}, the quality of computer simulation is validated by the comparison with the 
'ground truth'\cite{gt}
--
an a priori known true picture classification. 
But in our task the 'ground truth' {\em is} the internal human representation that we just want to simulate. Answering  the above question, we expect that the actual verification of our approach may be
mediated, as in the below example.


To generate the desired context,
we introduce a machine model for human perception called the {\em Alterego} (AE) net. 
In the course of this article, we first describe the approach to creating a 'universal' AE network. Then we consider how, given a data stream associated with a specific person, or a group of persons, 
to tune up the universal AE net to this person or the  group.

One may suggest various applications of these nets.
Imagine, for example, a group of several thousand people, where for every person from the group there exist an AE network
associated with the person. Suppose that prior to a championship in a certain city the networks are fed data (pictures, text) associated with the sporting event. 
Than if any network  generates a  content  semantically similar\cite{SemDist} to \textsl{"and then there were none"} song\cite{WereNone} (Agatha Christie\cite{Agatha Christie}), this   serves as a 'red flag' signal for the respective person intentions.

The contributions of this paper are as follows:
\begin{itemize}

    \item 
    The AlterEgo net is created {\em per person}.

    \item 
    
    It receives 
at input an object descriptions and produce at
output a representation of the internal person representation 
of the object, referred as the
    first level object representations.
    The first level object representation is 
    a flow of thoughts\cite{streamP} and feelings related
    to a perceived object.
   
    \item The BiWheel algorithm
    generates two interacting textual streams,
    similar to the narrative flow of 
    consciousness   \cite{streamN}.

    \item One of the above streams 
    is comprised of the word combinations
    semantically related to
    a perceived object. The stream
    is a second level representation of
    the human object representation. 
       
    \item 
    This second level representation is a
    description of the object which is
    not just a set of object properties,
    but the text stream.

\end{itemize}

Our paper is organized as follows.
We  start Sect.\,\ref{BiWheel} with modeling 
a flow of human thoughts\cite{streamP} and feelings by a sequence of iterations (Sect.\,\ref{SECTPAS}). 
Then extend the model to incorporate alterations of the perceived properties of 
objects (Sect.\,\ref{SECTOAS}) as iterations.
These two types of iterations are lump together in an iterative  scheme modeling human perception of objects (BiWheel scheme).
An algorithmic implementation of the scheme is 
presented in Sect.\,\ref{ImplBiWheel}. 
Further (Sect.\,\ref{Multiobject AE nets}), 
we describe how the introduced
AE network is generalized to perception of several objects.
Finally, we explain the personalization of the AE nets --  
tuning of the nets to a specific person or a group of persons, 
allowing generation of the representations reflecting 
their thought and felt contents, different for different humans
(Sect.\,\ref{Personalizion of AE nets}). 
Conclusions and description of the lateral
research directions associated with the AE nets complete the paper.

\section{The BiWheel scheme for modeling human object perception} 
\label{BiWheel}

\begin{figure}
\centering
\vspace*{-3.cm}
\includegraphics[scale=0.4]{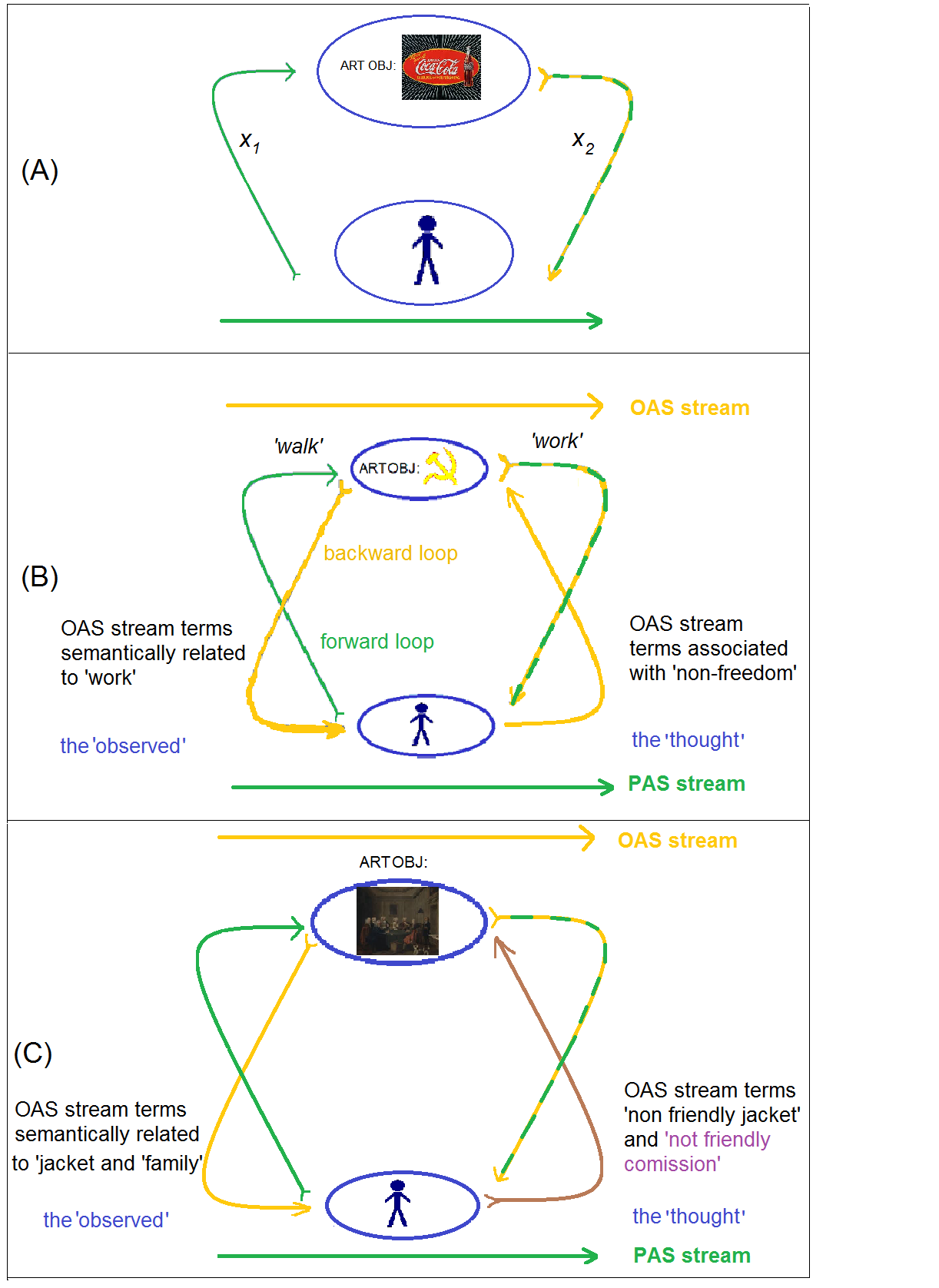}
\caption{
    Illustration to BiWheel scheme (best viewed  in color).
    The support and interleaving streams, are
    depicted respectively by green and yellow
    arrows.
    	 \newline  
    (A): Perception of the advertisement image (Sect.\,\ref{SECTPAS}).    
    \newline
    (B): Perception of the historical emblem
    \cite{hammer} (Sect.\,\ref{SECTPAS}).	 
    \newline
    (C): Perception of the social scene shown at 
    Fig.~\ref{fig:FIG14C}.	  
    \newline
    Brown arrows at (A), (B): the 
    elements inserted to
    the PAS (resp. OAS) stream may be obtained
    using external tools for
    image annotations. 
	}
\label{fig:BiWheel}
\end{figure}

In this section we describe our 
model of interactions of 
the flow 
of thoughts\cite{streamP} and feelings of a person
with the mental images
of the objects \cite{PerceptFarlex}.
The flow is modeled by 
textual streams resembling \cite{streamN}. 
The model is illustrated on the 
examples of an advertisement image, 
a historical emblem, and
a social scene.
We call the model the {\em BiWheel scheme}
because it includes two interacting loops  (Fig.\,\ref{fig:BiWheel} (B), (C)).

\begin{figure}
\centering
\includegraphics[scale=1]{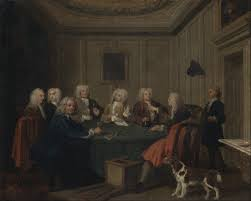}
\caption{
	The image from Fig.~\ref{fig:BiWheel} (C)\cite{vaada}.}
\label{fig:FIG14C}
\end{figure}

In Sect.\,\ref{SECTPAS}, 
we  describe a  'clockwise' loop 
comprising a first 'wheel' of  the scheme.
In Sect.\,\ref{SECTOAS}
the scheme is extended to modeling perception of 
the objects with
ambiguous meaning. This is done by
supplementing the scheme with
a 'counter-clockwise' loop
(Sect.\,\ref{SECTOAS}).
An overview of the scheme is given in Sect.\,\ref{compa}. 
In Sect.\,\ref{ImplBiWheel} we present the suggested algorithmic 
implementation of the BiWheel scheme.

\subsection{
Modeling a flow of human thoughts and feelings by a sequence of iterations} 
\label{SECTPAS}  

In this section we describe 
the first 'wheel' of the  
BiWheel scheme. 
We  consider two examples.
In the first one a person meets an advertisement image.
In a second example a person meets a historical emblem, or refer 
to it in his thoughts.

Consider a person walking on a road.
The person generates the flow of thoughts and feelings, reflecting its current mental state\cite{Mental state}.
The flow's elements at close temporal positions tend to be associated with each other
\cite{Association (psychology)}.

We may represent this flow as a sequence of words and short
word combinations with semantic meaning\cite{SemanticMeaning},
like: 
\begin{equation} \label{eq:PAS}
\textrm{PAS = 'street view', 'weather', 'cold', ...\,.}
\end{equation}
We  call such sequence a {\it Person Aligned Semantic } stream (PAS).
By some analogy with universal algebra operations
\cite{UniversalAlgebra} we  call the stream elements
the {\em terms}.
Since the PAS represents the flow of thoughts and feelings, the terms at closer positions tend to be closer semantically. 

\begin{figure}
\centering
\includegraphics[scale=0.35]{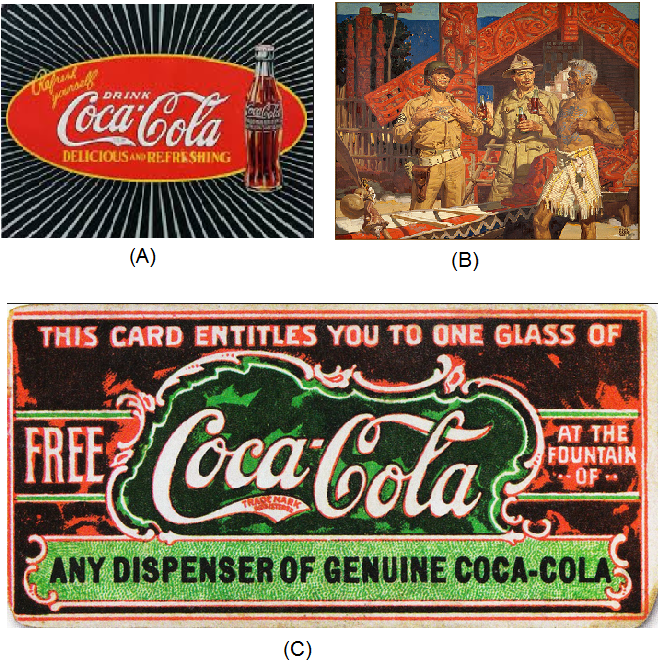}
\caption{
	The images
	refereed to in 
	Table\,\ref{cocacolatable}:
	(A), (B), and (C) are from
	\cite{Coca_wanna_drink},
	\cite{Coca_accomplishment},  and 
	\cite{Coca_aesthetic} respectively.
	}

\label{fig:FIG_COCACOLAS}
\end{figure}

Now suppose that during the walk the man  meets
an advertisement  image of Coca Cola (Fig.\,\ref{fig:FIG_COCACOLAS} (A)) which may influence
his flow of thoughts and feelings.
Let us ask whether there exist a linkage between the encountered image
and the man's thoughts and feelings,  {\em prior to} the act of image perception?
The answer is yes since the advertising targets the basic psychological needs
\cite{VainikkaBS}, in other words, the advertising
is virtually answering requests that already exist 
in the modeled flow of thoughts.
For example, among  the human's  thoughts there may exist the
represented in the PAS as:
\begin{equation} \label{eq:I wanna drink}
x_1 = \textrm{'I wanna drink'}   \in PAS
\end{equation}
The $x_1$ may be considered as 
a request virtually sending to the image
(Fig.\,\ref{fig:BiWheel}(A)).

Among the notions associated with the image content
there exist 
\begin{equation} \label{eq:Drink}
x_2 = \textrm{'Drink'},
\end{equation}
semantically related to $x_1$.
\footnote{
It is worth to note a certain duality\cite{Luure}: like the person's consciousness
before the act of perceiving the image  
contains  'requests' terms
associated with the image (like $x_1$), so the image 
 'contains' the 
terms,  which may be associated with the user's request (like $x_2$). 
This is a precondition
of the interaction between a person and an image.}

Transition $x_1\rightarrow x_2$ is illustrated at 
Fig.\,\ref{fig:BiWheel} (A): 
we submit to the object representation (the notions associated with the object)  an input, 
and obtain the response depending 
on the properties of the representation.
This resembles the forward step in 
in training convolutional neural networks\cite{forwardbackward},
which is basically the calculation of the network
value for a given input. Respectively,
we refer to such transitions as forward iterations. 

Forward iterations are comprised of 
'sending'
terms of the PAS stream to the object,
comparing them with the concepts\cite{concept}
associated with the  object, and
generating the 'answers',
sending back to the stream.
(The requests which are not relevant to the  object do not yield the responses.)
The iterations  may be seen as a 'dialogue'
between a PAS stream and
an object.

\begin{figure}
\centering
\includegraphics[scale=0.3]{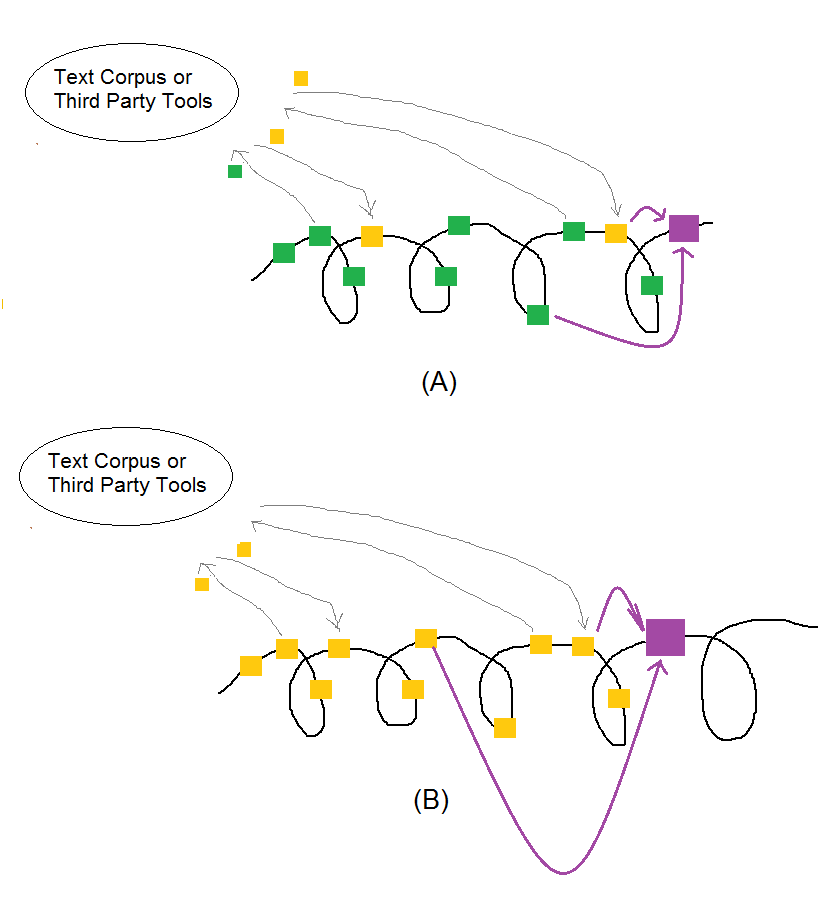}
\caption{
    (Best viewed  in color.)
    (A): The PAS stream. The 
    elements of the support substream
    are depicted in green, of the
    interleaving stream in yellow.
    (B): The OAS stream. The 
    elements of the support substream
    are depicted in yellow, of the
    interleaving stream in brown.
    Optionally, the elements of the
    interleaving streams 
    may be obtained by using 
    annotation tools.
    (A) and (B):
    In the both PAS and OAS streams
    the elements obtained by using
    selective summarization are depicted
    in violet.
}
\label{fig:PAS_OAS}
\end{figure}

The PAS can be seen now as comprised of the support and interleaving streams, 
as illustrated in Fig.\,\ref{fig:PAS_OAS} (A). 
The support stream consists of the terms, related to the 
person, like $x_1$ in Eq.\,\ref{eq:I wanna drink} and
sent as requests to the object.
The interleaving stream consists of the terms 
related to the object,
like $x_2$ of Eq.\,\ref{eq:Drink}, sending as
responses to the PAS.
By their meaning, the responses may conform
to the support stream (as in the above
example), or contradict it (as we will see in (Sect.\,\ref{SECTOAS}).
After the responses having been inserted to the PAS, they operate as regular stream elements, yielding the associated PAS terms.

\begin{table}[H]
\caption{Eventual interpretation of advertisement  images.}
\label{cocacolatable}
\begin{center}
    \begin{tabular}{ |  p{2.5cm} |  p{2.5cm} |  p{2.5cm} | p{2.5cm} |}
	  \hline
	  
     {\boldmath$x_1$}                & 
     {\boldmath$x_2$}                &
     {\bf The need category according to
     \cite{Maslow}
     
      }               &
     {\bf Image}
	    \\ \hline

	'I wanna drink' 		    & 
	'Drink'   	&	
	Physiological 	       	&	
	Fig.\,\ref{fig:FIG_COCACOLAS} (A)
	\\ \hline

	Feeling of accomplishment	    & 
	Feeling of accomplishment   	&	
	Self-Actualization       	&	
	Fig.\,\ref{fig:FIG_COCACOLAS} (B)
	\\ \hline

	An aesthetic feeling    & 
	An aesthetic feeling   	&	
	Aesthetic        	&	
	Fig.\,\ref{fig:FIG_COCACOLAS} (C)

    \end{tabular}
\end{center}
\end{table}

In our second example we show how
the forward iterations 
model perception of historical emblems. 
Let us consider Fig.~\ref{fig:BiWheel} (B), where hammer and sickle \cite{hammer} are depicted. As in the above example, consider a man walking on a road.
Let us imagine 
the following  'dialog' 
between the PAS stream and the set of 
the concepts
associated with the emblem.

In the PAS stream (Eq.\,\ref{eq:PAS}), at some iteration step, is generated a term 
\begin{equation} \label{eq:walk}
x_1 = \textrm{'I wanna go walk'}.
\end{equation}
Similarly to the above example,
the term is compared with the ones associated with the emblem. There may be found a term 
\begin{equation} \label{eq:work}
x_2= \textrm{'You are going to collective work to help the motherland!'}
\end{equation}
semantically related to $x_1$,
depicted with green dashed lines in Fig.~\ref{fig:BiWheel} (B).
As in the example with the advertisement image, 
$x_2$ may be inserted to the PAS, and
and the forward iterations will continue.
On the other hand, the $x_2$ may 
fire a contradiction signal with the PAS (like neuron \cite{Artificial neuron}) and thus not be inserted 
to the stream. This option is considered in the 
next section.

In the examples considered in this section 
the perceived object 'operates' like a  function: 
it receives an input term $x_1$ 
and returns the response. 
In the following section we consider how 
alterations of  perceived {\em properties} of 
objects may be modeled.

\subsection{Modeling alterations of perceived object's properties
\label{SECTOAS}} 

When we look at advertising images,  their meaning is unequivocal --
they rarely have ambiguous interpretation. For many others
perceived objects it is not so. For example, historical emblems may have quite different meanings. For instance, the 'Hammer and sickle' symbol
of the former Soviet Union \cite{SU} was associated by some people with the ideas of  freedom \cite{commi, freedom},
and on the other hand, by other people, their negation\cite{gulag}. 
Another example of objects that allow different interpretations are social scenes and situations estimated by persons. 
The meaning of such objects may be altered during the perception process \cite{Percept}.

In this section we  extend 
the PAS model of Sect.\,\ref{SECTPAS} -- 
it will simulate also alterations of perceived 
properties of objects during the perception act.
This is done
by adding the second
iteration loop to the BiWheel model.
The model extension will be illustrated on the examples of
a historical emblem and a social scene.

Let us consider a sequence of semantically
related terms associated
with a perceived object. 
We call it the {\em Object Aligned Semantic} (OAS) stream. This stream consists of the terms describing 
perceived properties of the object. 
Actually we have dealt with this stream in the previous section, this is the stream where from the terms $x_2$ were selected. 
For example, for the image from 
Fig.\,\ref{fig:FIG_COCACOLAS} (A) the OAS may consist
of the terms like
\begin{equation} \label{eq:OAS}
\textrm{OAS ='red', 'Coca Cola', 'drink', 
'star','rays',...\,.}
\end{equation}
It may be regarded as 
a stream of annotations of the advertising
image.

Let us turn 
again to the second example from the previous section.
What will happen if the ideas\cite{Idea} 
associated with the emblem  come 
in collision with the human thoughts and feeling?
Then likely the person's thoughts will be switched to
the object itself, leading to rethinking  its properties. 
As a result the meaning of the 
object for that person may be alternated.

In our model this is represented as 
firing collision between the object representation's response 
($x_2$ Eq.\,\ref{eq:work})
and the notions comprising the PAS stream of the person.
It leads to switching the data processing to the OAS stream, 
and starting enumeration of the terms semantically close to $x_2$. 
This  might lead to a change in the semantic 
meanings (currently) associated with the object.

Our next example is illustrated by Fig.~\ref{fig:BiWheel} (C)
and related to estimation of social scenes. 
Suppose a person is summoned to a commission
(illustrated at Fig.~\ref{fig:FIG14C}),
which will make a decision regarding him, and it is known in advance that the decision may be biased, for the person or against.
\footnote{The example was developed basing on some person's experience at the oral entrance examination for Moscow University Faculty of Mechanics and Mathematics in the past.}
The person does not know the bias and hesitates.

As in Sect.\,\ref{SECTPAS}
we  model the
person's flow of thoughts as a PAS stream of terms, say:

\begin{equation}
\begin{array}{l }
\textrm{PAS = 'they', 'they are kind to us',
'this city light is friendly to us',}\\

\textrm{ 
$\,\,$ 'they', 'they are kind to us', 'this city light is friendly to us',} \\

\textrm{
$\,\,\,$'look, they have the same jacket as we have',} \\

\textrm{
$\,\,\,$'this is a deeply friendly jacket', ...} \\

\end{array}
\end{equation}  

Similarly to the above example (Eq.\,\ref{eq:OAS}), 
we model the mental image of the environment (commission) by an OAS stream:
\begin{equation} \label{eq:OAS2_0}
\begin{array}{l }
\textrm{OAS = 
'jacket', 'commission',
'people', 'came against us',  
}\\

\textrm{ 
$\,\,$  
'came for us', 'does not matter',
...
} \\

\end{array}
\end{equation} 

At some point in time,  terms 
$x_1$ from the PAS are compared with the environment -- verified w.r.t. the  mental  image of the
commission and initiate the responses
$x_2$. 
The obtained response 
may contradict to the PAS stream, e.g.,
for $x_1 = \textrm{'this is a deeply friendly jacket'}$, 
the $x_2 = \textrm{'this jacket is not friendly to us at all'}$.
At this moment the forward iterations are
interrupted, and our simulation process is switched to
enumeration of the properties of the commission,  starting with $x_2$. The iterations yield a new portion of
an OAS stream, like: 
\begin{equation} \label{eq:OAS2}
\begin{array}{l }
\textrm{OAS = 
...
'this jacket is not friendly to us at all',
'jacket','commission',}\\

\textrm{ 
$\,\,$  'people', 'came against us',  'came for us', 'the jacket does not matter',} \\

\textrm{
$\,\,\,$'they want to frighten us with their uniform', 'the hostile unity of their } \\

\textrm{
$\,\,\,$'clothes', ...} \\

\end{array}
\end{equation} 

In the both above examples, the elements $x_2$ from the OAS
serve as magnets attracting the person's attention,
influencing formation of subsequent OAS
elements. Accordingly,
these elements may be seen as forming an interleaving stream 
to the support (proper) OAS. 
In this way, similarly to the PAS, the OAS stream 
is comprised of the support and interleaving streams. 

If we interpret obtaining the
'answers' $x_2$ from the OAS stream to $x_1$
as a calculation of the value of an OAS 'function' at  $x_1$, 
then generation of the new terms
of Eq.\,\ref{eq:OAS2}
may be treated as updating the function 
parameters. This resembles
the backpropagation step in training convolutional neural networks \cite{forwardbackward},
where the network parameters are updated.   
We  keep this notation 
for the whole iteration loop (Figs.~\ref{fig:BiWheel} (B), (C))
together with the forward iteration notation
(Sect.\,\ref{SECTPAS}).

In forming the OAS streams
there may occur what may be called {\em selective summarization} - selection of a small set of the terms consistent with the  stream.
For example, 'non friendly jacket' and 'my predecessor has suffered from the commission' appeared in
the OAS of Eq.\,\ref{eq:OAS2},
may 
lead to a new term 'non friendly commission for me' (Fig.~\ref{fig:PAS_OAS} (B)).
This may be seen as a kind of compression
of the OAS. 
Similar summarization may act on the PAS stream.

It is interesting to note  
that forming the PAS and OAS
streams  resembles
exploring the YouTube
\cite{YouTube Wiki} 
engine. 
Indeed, each stream
looks like a sequence of annotations  
displayed  by the engine after a user has picked a video.
We  explore this similarity 
in Sect.\,\ref{ImplBiWheel}.

In the latter two  sections we have studied
the forward and backward loops comprising
the BiWheel algorithm. In the next section we overview the whole scheme.

\subsection{Overview of BiWheel scheme. Received and perceived. }
\label{compa}

Generation of the PAS and OAS streams  is
based on the following 
principles:

\begin{enumerate}

\item
 
In each stream the generated terms tend to be semantically close to 
recently generated terms of the stream.

\item

Periodically, the generated terms of each stream tend to be 
semantically related to terms of the opposite stream. 

\item

Some terms of the OAS stream may be input from external annotations tools. 

\end{enumerate}

One may summarize the work of the BiWheel scheme as follows
(Fig.\,\ref{fig:OVERALL}):

\begin{figure}
\centering
\includegraphics[scale=0.4]{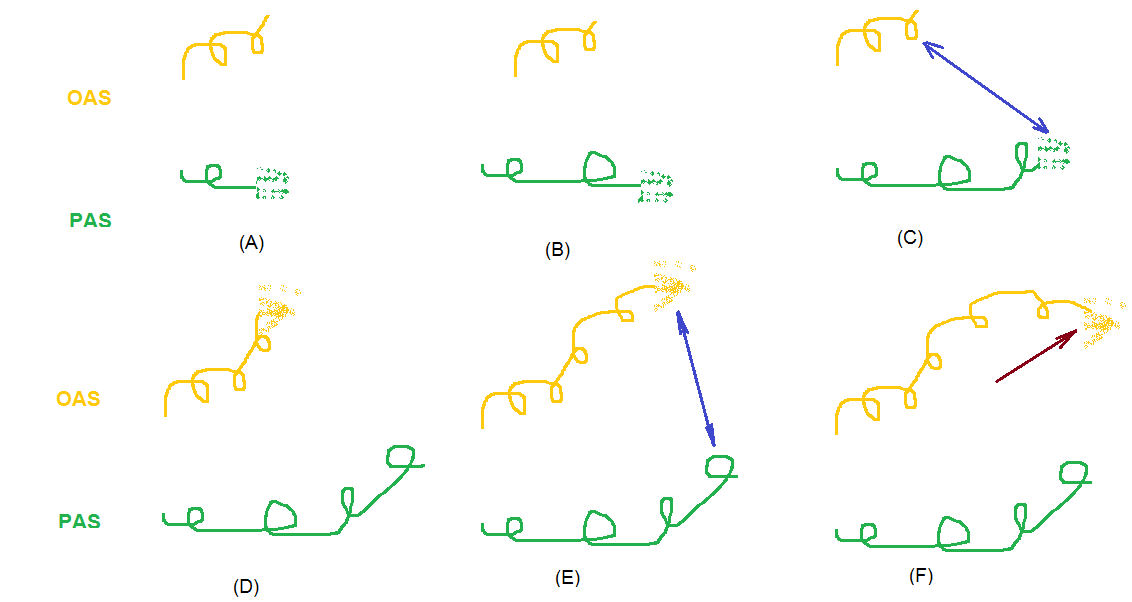}
\caption{
(A)--(E):Sequential forming the PAS and OAS streams by BiWheel scheme. The added elements are denoted by dots.
(A),(B),(C): new element are added to the PAS stream are semantically related to the previous PAS elements. 
(C): The new element is semantically related to the terms from the OAS (considered as the element from the interleaving stream).
(D),(E),(F): new elements are added to the OAS stream are semantically related to the previous OAS elements. 
(E): The new element is semantically related to the terms from the PAS (considered as the element from the interleaving stream).
(F): The new element is obtained from the annotation tool (considered as the element from the interleaving stream).
}
\label{fig:OVERALL}
\end{figure}

\begin{enumerate}

  \item PAS is a stream of mutually associated terms representing
   the alternated mental state of a person. 
  \item OAS is a stream of mutually associated terms representing the alternated 
  perceived properties of the object.
  \item Current elements of the PAS stream are 
   compared with the semantically related current OAS
   elements, the latter are sent to the PAS as the
   OAS responses. 
  \item These responses are embedded to the PAS stream, 
  or (if  contradict to PAS) invoke
  a reestimation of the object properties.
  \item The reestimation  proceeds as adding to the  OAS    
  new terms associated with the responses, or adding
  the object's descriptions from annotation tools.
  \item The summarization is optionally performed on 
  both the streams.
\end{enumerate}

\begin{figure}
    \centering
    \includegraphics[scale=.55]{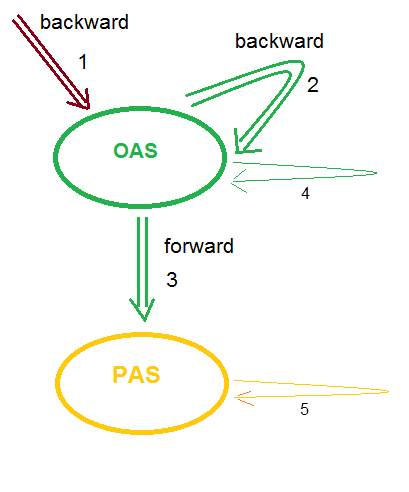}
    \caption{Asymmetric data transfer between the PAS and OAS streams.
    The arrows 1, 2, and 3 are explained in the text.
    The arrows 4 and 5 denote the support PAS and OAS streams
    (Sects.\,\ref{SECTPAS}\,and \ref{SECTOAS}\,).
    }
\label{fig:FIG_PAS_OAS_NONSYM}
\end{figure}

Let us note that each of the generated streams may  include term repetitions.
It is worth to mention that the interaction 
between the PAS and OAS streams is non symmetric. 
As we saw above, at forward iterations, elements of
OAS are embedded to PAS and start to participate
in forming 
the stream (arrow 3 in Fig.\,\ref{fig:FIG_PAS_OAS_NONSYM}). 
In contrast to this, at backward iterations, 
no new elements are embedded to the OAS stream.
Namely, at the start of a backward iteration,
the elements of OAS that have invoked switching 
from the forward to backward iteration, are beginning to form  
continuation of the stream (ibid, arrow 2). 
Together with annotations from external tools 
(ibid, arrow 1), the asymmetry indicates the transfer 
of a new 'information' from outside the streams 
to the OAS,	and then to the PAS stream. 

The  OAS stream 
is constructed in such a way that 
it depends both 
on the perceived object and on the person. 
Indeed, this is the {\em object} describing content, because the OAS is generated 
per object. 
And this is 
{\em person}  derived content  because
the OAS is depends on interactions with the PAS 
stream, which reflects the internal world of the person.
But how the OAS stream approximates the 
{\em perceived} content (Fig.\ref{fig:FIG_THOUGHTS_PERCEPT} (D), (E))? 

An OAS stream 
starts to be generated being not dependent on the person (Eq.\,\ref{eq:OAS2_0}): these streams built for different persons may 
begin with the same terms.
Furthermore, 
the elements of the OAS  stream
embedded to the PAS (Sect.\,\ref{SECTPAS}) may be interpreted as 
perceived object properties.  
These may be seen as the OAS terms the person 
associates with the current PAS terms. 
Although their selection depends on the person,
it does not affect the OAS stream. 
Furthermore,
the {\em interleaving} OAS stream 
is comprised of the OAS terms
which have {\em switched} the person's attention,
and so convey a person's 'imprint'.
Thus the interleaving stream  may be seen as consisting of 
the perceived object properties
in a strong sense.
Furthermore, the interleaving stream influences the whole
OAS stream.

In this way 
the OAS stream 
is a representation of the
'thought \& felt' 
content of the person,
whereas the
interleaving substream 
may be considered as representing 
this content in a strict sense. 

It is interesting to note that the {\em work} of the  scheme is reminiscent of {\em training} the neural network\cite{CNN}. Indeed, the former is comprised of iterations that may be referred as  forward and backward (Sects.\,\ref{SECTPAS},\ref{SECTOAS}). 
Thus, our
modeling of human perception may be considered as reinforcement learning\cite{Reinforcement learning} of the network associated with the perceived object.
For example, perception by a person of an apple is modeled  as the person's dialog with a network 'Apple', accompanying with a simultaneous training of the network. 

The properties of the BiWheel loops 
are summarized below in Table \ref{commontable} in
informal fashion.

\begin{table}[H]
\caption{Informal properties of the dual loops: comparison.}
\label{commontable}
\begin{center}
    \begin{tabular}{ |  p{4cm} |  p{3cm} |  p{3cm} | }
	  \hline
	  
     {\bf Activity:}               & 
     {\bf Forward}                 &
     {\bf Backward}
	    \\ \hline

	The notation associated: 		& 
	internal, inhale, passive, atomistic   	&	
	external, exhale, active, holistic								           
	    \\ \hline

	Direction: 						& 
	Clockwise 						&	
	Counter-clockwise								
	    \\ \hline
	    
    Stream name:                    &        	    
    Person Aligned Stream  = 
    support + interleaving stream   &
    Object Aligned Stream = support + interleaving stream
	    
	    \\ \hline	    
	    
    Support stream:                  &
    A sequence of associations yielded 
    by the person's paradigm

    &
    A sequence of associations 
    yielded by the object's representation
    of a human

	    \\ \hline	    
	    
    Interleaving stream: 						&
    Responses from the object to the support stream
    based on 
    associations of the terms

           				&
    Insertions to the
    support stream based on 
    associations of the terms, or
    obtained using  from the external annotation
    tools
	    
	    \\ \hline

    Interpretation of the interleaving stream:                   &
	            	    
    The 'thought \& felt by the object '
                                                    &
    The 'thought  \& by the person'

    \end{tabular}
\end{center}
\end{table}

\subsection{Implementation of BiWheel Scheme}
\label{ImplBiWheel}

Below we  describe the suggested simplest algorithmic implementation of 
the scheme.
It simulates a perception of an 'average' person,
located in an environment $ENV$ of an object $OBJ$.

Both the PAS and OAS streams are comprised
of the terms of a text corpus. 
We initialize the PAS to a small family 
\begin{equation} \label{eq:W}
\overline{W} = \{\overline{w_1},\overline{w_2}, \dots , \overline{w_I}\}
\end{equation}
of the words and short word combinations describing the environment
$ENV$ of a person. For example 'street', 'going along', 'the weather in the city', etc. We refer to those terms as the {\em generating elements} of the  stream. Our goal is to generate the PAS 
(Eq.\,\ref{eq:pasterms})
subjected to the following conditions:

\begin{itemize}
\item Environmental condition: terms $w_i$ are semantically close to 
the terms from $\overline{W}$;
\item Continuity condition: terms $w_i$ tend to be close to recently generated terms
$\{w_{i_0} \,|\,  i_0 <i\}$.

\end{itemize}

We  denote the PAS stream 
generated subjected to these conditions by:
\begin{equation} \label{eq:PAS2}
PAS \, = \,\, < \overline{W} >.
\end{equation}

The stream is constructed as a sequence of terms  
    \begin{equation} \label{eq:pasterms}
    PAS = w_1,w_2,\ldots ,
    \end{equation}
in such a way that at every step $i$
a new term $w_i$
is generated subjected to 
the environmental
condition
with probability $p$, and to the continuity condition
with probability $1-p$, where $p$ is a predefined probability value.
In the former case $w_i$ is constructed 
to be similar to at least one 
term from $\overline{W}$.
In the latter, it is constructed 
to be similar to the term, randomly selected
(with probability proportional to $\lambda$ weight assignment)  from the set of
recently 
generated terms (Algorithm\,\ref{algo:BiWheel}).
In both the cases, $w_i$ is selected 
to be semantically similar to 
a certain term 
$w' \in \{ w_j \,| \, j < i\}$, 
referred further as attractor.


Given an attractor $w'$, construction of $w_i$ may proceed 
by retrieval terms from text corpora and selection those semantically close to $w'$. 
For example, to treat the
attractor as a document 
and map it to a set of documents with 
semantically similar content 
using semantic hashing\cite{Semantic Hashing Blog}.
Then one may perform 
an exhaustive search of n-grams
\cite{n-gram} in the retrieved documents,
and select the terms semantically similar
to the attractor \cite{Using N-gram}.

Let us notice
that the generation of 
the PAS stream (Eq.\,\ref{eq:pasterms}) resembles retrieval
operations using
the YouTube
\cite{YouTube Wiki} 
engine, where a user randomly 
searches the items related to the 
input terms
$\overline{W}$. 
Indeed, fulfillment of the environmental condition is
similar to queering the 
engine with terms from $\overline{W}$
and assigning $w_i$ the annotation 
of a video $vid$
randomly selected from the list of retrieved videos.
Respectively, fulfillment of
the continuity condition is similar 
to selecting  $vid$,
and assigning to a $w_i$ the annotation 
of a randomly selected video 
from the YouTube suggestion list \cite{suggestion youtube}.
In such a way the PAS stream could be 
generated using deep neural networks, 
like the YouTube recommendations algorithm\cite{YouTube}.

Likewise to PAS, we initialize the OAS stream
to a family of terms 
describing the properties of the input object $OBJ$:
\begin{equation} \label{eq:V}
\overline{V} = \{\overline{v_1},\overline{v_2}, \dots , \overline{v_I}\}.
\end{equation}

For example, if the described object is an apple, then
\begin{equation}
\overline{V}(Apple) = \{
\textrm{apple},
\textrm{green apple}, 
\textrm{fruit},
\textrm{fruit nutrition} 
\dots\}.
\end{equation}

Similarly to PAS, we may generate the 
OAS subjective to the environmental (proximity of  
the generated terms $v_i$ to 
$\overline{V}$ ) and continuity conditions, denote it as:

\begin{equation} \label{eq:OAS2_2}
OAS \, = \,\, < \overline{V} >.
\end{equation}

%

The work of our BiWherel scheme is
summarized in Algorithm\,\ref{algo:BiWheel}.

\begin{algorithm}
	\caption{The simplest AE net
	\newline
	{\bf input}: 
	environment {\em ENV},
	perceived object {\em OBJ}
	\newline
	{\bf output}: OAS stream
	}
	\begin{enumerate}
	
		\item \label{itm:init}
		Initialization:
    	\newline
            Generate the starting portion $P$ of 
		    PAS stream, 
		    $P=\overline{W}$,
		    following Eq.\,\ref{eq:W}.
        	\newline
            Generate the starting portion $O$ of 
		    OAS stream,
		    $O=\overline{V}$,
		    following Eq.\,\ref{eq:V}.


		\item \label{itm:controlller}
    		Controller:
        	\newline
    		perform  
	    	Step \ref{itm:pas},
		    Step \ref{itm:oas},
    		Step \ref{itm:pas},
	    	Step \ref{itm:oas} \ldots

		\item \label{itm:pas} Forward iteration:
		
            \begin{enumerate}[label*=\arabic*.]		
		
    		    \item  \label{itm:pas new portion}

        		    Form a new portion $P$ of 
	        	    PAS stream, 
		            generating the
		            elements $w_i$
		            semantically close
                    to ${ \overline W}$ and 
        		    to the element with
	        	    larger weight $\lambda$

    		        After the portion generation,
	    	        assign 
    	    	    weights $\lambda(w_i)=1$, 
    		        to the recently generated
    		        elements, 
    		        normalize distribution
    		        $\lambda$ to unit sum.

		        \item \label{itm:pas-find-oas} 
        		    For PAS elements of $P$ 
        		    find a set $S \subseteq O$ of semantically
        		    related
        		    elements in OAS, preferring  
        		    the elements with larger 
        		    weight $\mu$.
        		    
                    \begin{enumerate}[label*=\arabic*.]		
        		    
    		        \item \label{itm:pas-find-oas-basic} 
        		    Construct 
        		    $S_1  \subseteq S$, 
        		    consisting of 
        		    elements semantically similar
        		    to the PAS elements. 
        		    \newline
        		    Form interleaving portion to PAS:
        		    add the terms from 
        		    $S_1$ to  the PAS with 
        		    weights $\lambda=1$, 
        		    normalize distribution
        		    $\lambda$ to unit sum.
    		        \item \label{itm:pas-find-oas-interl} 
        		    Construct 
        		    $S_2  \subseteq S$, 
        		    consisting of 
        		    elements semantically contradicting
        		    the PAS elements.
        		    \newline
        		    Form interleaving portion to OAS: 
					add $S_2$ to  the OAS with 
                    weight $\mu=1$,
        		    normalize distribution
    		        $\mu$ to unit sum.
                    
                    \end{enumerate}                    
                    
		        \item \label{pas ret} 		    
            		Return to the controller block.
		    
            \end{enumerate}

		\item \label{itm:oas}  Backward iteration:
		
            \begin{enumerate}[label*=\arabic*.]		
            
                \item \label{oas O}
		   		    Form a new portion $O$ of 
	        	    OAS stream, 
		            generating the
		            elements $v_j$
		            semantically close
                    to ${ \overline V}$ and 
        		    to the element with
	        	    larger weight $\mu$
%
		    
    		        After the portion generation,
	    	        assign 
    	    	    weights $\mu(w_i)=1$, 
    		        to the recently generated
    		        elements,normalize distribution
    		        $\mu$ to unit sum.

                \item \label{itm:third} 
		            Optional step.
        		    Add new elements to the 
	        	    OAS using the 
        		    annotation tools.
    		        After the portion generation,
	    	        assign 
    	    	    weights $\mu(w_i)=1$, 
    		        to the recently generated
    		        elements,normalize distribution
    		        $\mu$ to unit sum.
		    
		        \item \label{oas ret} 		    
            		Return to the controller block.
		    
            \end{enumerate}
		    
	\end{enumerate}
\label{algo:BiWheel}	
\end{algorithm}

Over the course of the algorithm running, the set of the terms $S$ in Step.\,\ref{itm:pas-find-oas} is constructed by enumeration of 
the pairs of terms 
\begin{equation} \label{eq:L}
L = \{ \, (t_1,t_2) \} , t_1\in P, t_2\in O \, \},
\end{equation}
selecting  
the set of $L_1 \subset L$ of semantically related pairs of the terms, 
and calculation of
projection 
\begin{equation} \label{eq:S}
S=\{ t_1 \,|\, (t_1,t_2) \in L_1\}.
\end{equation}
The $S_2  \subseteq S$ in Step.\,\ref{itm:pas-find-oas-interl} 
is constructed by selecting
$L_2 \subset L_1$ consisting of semantically contradicting
terms $(t_1,t_2) \in L_1$ \cite{semantically contradict},
following projection calculation:
\begin{equation} \label{eq:S2}
S_2=\{ t_1 \,|\, (t_1,t_2) \in L_2\}.\end{equation}

Our description of the BiWheelscheme 
given in (Sect.\,\ref{BiWheel}) may cause the question:
why AE is called the {\em  net}? 
The AE implementation considered in this 
section answers this question.
Indeed, the AE is implemented using 
a kit of NLP instruments which, in their turn,
are naturally implemented using the deep learning 
tools. Therefore, the AE may be seen as 
a compound deep learning net.

The networks AE considered up to this point
are limited by two main factors:
networks model the perception of one object and do not imitate 
human perception  in a {\em person-dependent fashion}.
In the next section, we extend the BiWheel 
model to overcome these limitations.

\section{AE nets for simulation of human perception} 
\label{AE nets for simulation of human perception}

In the previous sections we described the 
simplest AE net which consists of a single BiWheel scheme. 
The net models perception of a single object. 
In this section we  introduce
the Multiobject AE nets allowing to model 
perception of several objects. 
We  also discuss
how Multiobject AE nets
may be tuned for 
modeling perception of different persons.

    \subsection{Multiobject AE nets} 
    \label{Multiobject AE nets}

\begin{figure}
    \centering
    \includegraphics[scale=.3]{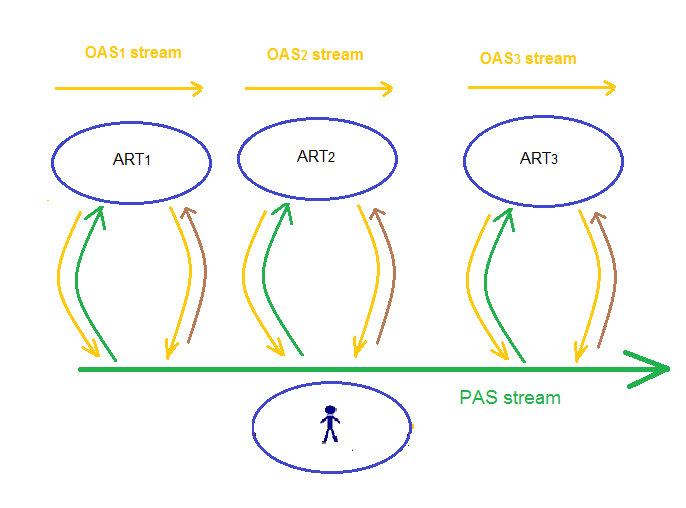}
    \caption{Multiobject AE net. }
\label{fig:FIG_MULTIOBJ}
\end{figure}

Multiobject AE net
is generalization of the BiWheel scheme 
for simulating perception of several objects. 
Given a BiWheel scheme and a family of 
a new objects the Multiobject AE
is constructed 
by incorporation to the scheme 
the OAS loops,
representing the objects. 
As the result, the PAS stream of the new net
is receiving the interleaving embedments from several OAS streams. This is illustrated at Fig.\,\ref{fig:FIG_MULTIOBJ}.  

One may notice that our implementation of the 
simplest AE net (Sect.\ref{ImplBiWheel}) is easily generalized to the multiobject 
format. 
The Multiobject AE may be implemented in multithread fashion, one thread per object, in such a way that every thread is responsible for forming the OAS stream for the respective object.

    \subsection{Personalizion of AE nets} 
    \label{Personalizion of AE nets}

The AE nets considered so far
do not deal with a person
specific information and  accordingly
model perception of an 'average' human. 
Tuning to the specific person may be
done incorporation to the input environment 
of the scheme (Sect.\,\ref{ImplBiWheel})   
the person specific information.
Given a person 
$H$ and a set of terms $T(H)$ representing this information
(this may be referred as the set of person specific features terms), we may add $T(H)$ to the generating set of the PAS stream (Eq.\,\ref{eq:PAS2}):

\begin{equation} \label{eq:PAS_H}
PAS_H \, = \,\, < W \cup T_H>.
\end{equation}

\begin{figure}
    \centering
    \includegraphics[scale=.3]{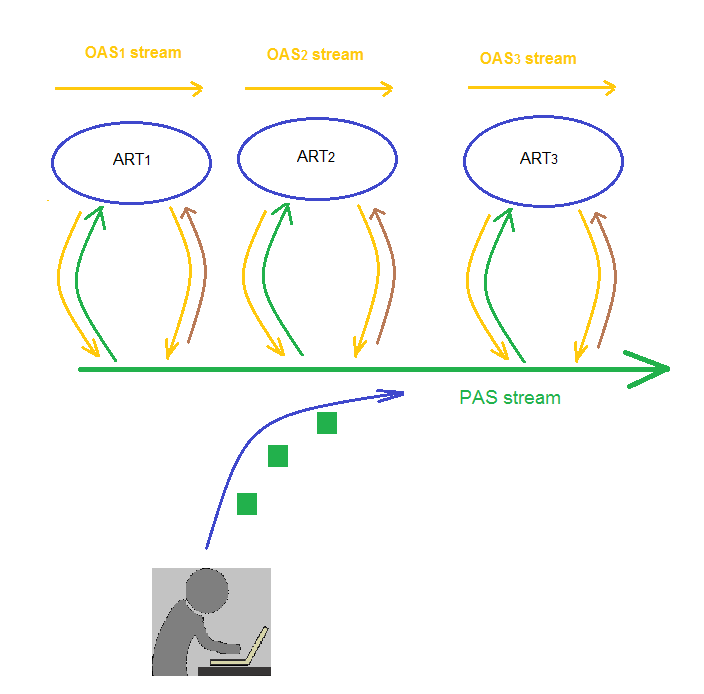}
    \caption{Personalized AE net.}
\label{fig:FIG_PERSONALIZATION}
\end{figure}

The $T_H$ may be constructed as a verbal description of the person, e.g.,  

\begin{equation} \label{eq:T_H}
T_H \, = \,\, \{  \textrm{'age: 35 -- 50'}, \textrm{movie:"Call Me by Your Name"}   \},
\end{equation}
or be retrieved from  the content environment associated with the the person, like the search engine  queries. 
This is illustrated at Fig.\,\ref{fig:FIG_PERSONALIZATION}.
In a similar fashion, 
the personalization may be performed for a group of of persons. 

\section {Conclusions and lateral research directions}

In this paper we introduced a scheme  
for modeling internal 
human representation of perceived objects.
The  scheme is based on two pivots. First is the idea that 
ability to dialog characterizes a human\cite{MeditationsOfTheRebbe}. 
Second, the inference,
inspired by recent advances in NLP, like Word2Vec, 
claiming that iterations may be more important than their non-iterative 
counterparts\cite{iterat}. 
We simulate the object perception as 
iterative dialog between the persons.
We believe that this simulation may be more powerful than  
descriptions 
comprised of the 'static' information.

The scheme generates the data  
approximating the internal human content.
This allows to suggests  numerous scenarios for its application. 
For example, given conglomeration of persons, one may associate with every person of the conglomeration an AE net, and train
the nets using the observations associated with 
the persons. Then for a certain input object,
for example, an advertisement  image,
the trained nets 
 generate 
the content approximating how
the people perceive the object.
This may be used for construction of
the input objects that are desirable by the persons, in {\em  per person}
fashion. 
In another scenario, the AE nets 
associated with every person from
a given conglomeration are used 
for search of the people with a specified characteristics of their perceived content.
An example of such application has been considered in the introduction.
In a third scenario, 
the AE nets may be  generalized to
simulate {\em communication} between 
the contents perceived by different people. 
In such a way one may model a {\em crowd perception}
\cite{Crowd psychology}
\footnote{Of course, such social measurements can only be carried out under control provided with social mechanisms.}.

\begin{figure}
    \centering
    \includegraphics[scale=.15]{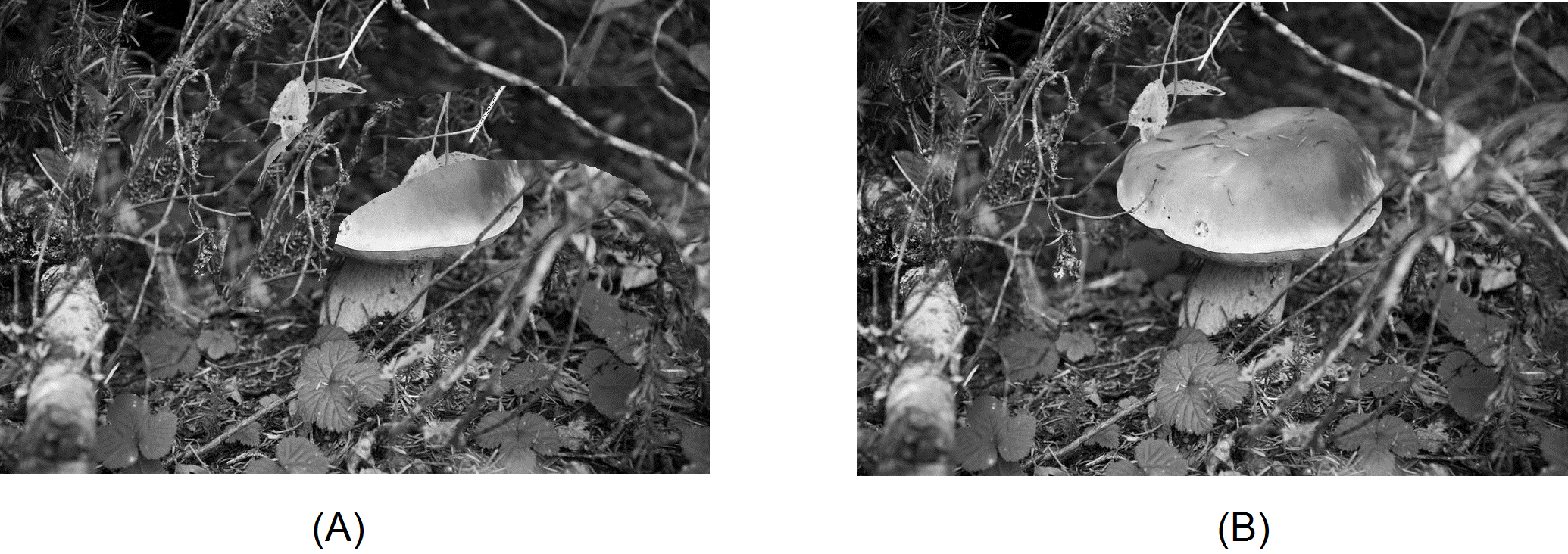}
    \caption{
		Whether we may train a neural network
		to seek the mushrooms in the photographies, and
		given an input image (A)
		\cite{mushr_changed},
		to yield the  image 
		similar to (B)
		\cite{mushr_orig}? 
    }
\label{fig:FIG1AB}
\end{figure}

There are exist additional directions of the research of AE. The first is related to 
modeling the percfeived {\em visual}
content. The
AE nets described in this paper 
model the human {\em thought} content
by textual stream. 
It may be desirable 
given a 
perceived object
(Fig.\,\ref{fig:FIG_THOUGHTS_PERCEPT} (A)),
to model the
mental image\cite{PerceptFarlex}
(Fig.\,\ref{fig:FIG_THOUGHTS_PERCEPT} (C))
by digital image
\cite{Digital image}
(ibid (E)), i.e., to model the {\em 'seen'} instead of the {\em 'thought'}.
It is worth to note that from the point of view of the artificial intelligence research, modeling the 'seen' 
is not bound to simulation of the human perception. 
Indeed, it may be desirable to enable the neural network
trained to seek the mushrooms in the photographies,
given the image shown Fig.\,\ref{fig:FIG1AB} (A),
to yield the output similar to that of 
ibid (B). 
This may resemble the Deep Dream 
approach\cite{DeepDream}, but in contrast to the latter, 
the sought image should
be obtained at the {\em output}
end  of the AE network, and not at the  end of the input image.
This is the topic of our current research.

Another lateral direction of the AE research is their use 
for the human content {\em creation}, for example, automatic poetry generation\cite{Automatic generation of poetry}.
We outline this approach in Appendix
\ref{appendix A}.

We believe that AE networks will find applications beyond the described in this article.

The image and text data 
from\,
\cite{commonDQ}\,--\cite{Coca_aesthetic}
were used in the preparation.

\section*{Acknowledgment} 
\addcontentsline{toc}{section}{Acknowledgment}

I wish to acknowledge Drs. 
Eyal Nitzany,
Daniel Segel, and Michael Katz for their help that greatly improved the manuscript.

\appendix
\section{Use of BiWheel scheme for the automatic generation of the art content}
\label{appendix A}  

It seems very likely that the BiWheel scheme may be used also for simulation 
of the art content {\em creation}, for example, automatic poetry 
generation\cite{Automatic generation of poetry}.
Let us show how the scheme is related to
to the perception of the {\em existing } poetic texts. Consider the following example.
A human may obtain aesthetic impression while reading 
a piece of poetry describing the sea (e.g., \cite{Mand}), if  suddenly feels that the 
sea's noise referred in the poem
resembles 
the physical rhythm of the text read. 
The "sea's noise" 
may be seen as elements
of the PAS stream 
(Sect.\,\ref{SECTPAS})
representing 
the human flow of thoughts 
generated while perceiving the poem.
The rhythm of the text
may be associated with the subsequence of the OAS stream 
(Sect.\,\ref{SECTOAS}) describing 
perceived physical properties
of the text (along with visual text appearance). 
Receiving the impression of resemblance mentioned above,
may be seen as firing similarity signal between 
the PAS and OAS streams.
In such a way, the peaks of aesthetic impression correspond to the 
pairs of semantically similar elements in these streams (depicted in Fig.\,\ref{fig:OVERALL}
by blue segments).
Analogously, our model may describe perception of 
aesthetically  efficient texts of small length, like "Rain, Steam and Speed" of  J. M. W. 
Turner\cite{RSS},
or presented in\cite{CG}.
It seems therefore that our scheme may be used is similar fashion as {\em generative}
model for the art content creation.

Probably, similar mechanism may be used also for automatic generation
of the simplest mathematical propositions, for example,  
from the groups theory\cite{groups}.



\end{document}